# Delivery Issues Identification from Customer Feedback Data

Ankush Chopra
Tredence Analytics Sol. Pvt. Ltd.
ankush.chopra@tredence.com

Mahima Arora
Tredence Analytics Sol. Pvt. Ltd.
mahima.arora@tredence.com

Shubham Pandey
Tredence Analytics Sol. Pvt. Ltd.
shubham.pandey@tredence.com

*Abstract*—Millions of packages are delivered successfully by online and local retail stores across the world every day. The proper delivery of packages is needed to ensure high customer satisfaction and repeat purchases. These deliveries suffer various problems despite the best efforts from the stores. These issues happen not only due to the large volume and high demand for low turnaround time but also due to mechanical operations and natural factors. These issues range from receiving wrong items in the package to delayed shipment to damaged packages because of mishandling during transportation. Finding solutions to various delivery issues faced by both sending and receiving parties plays a vital role in increasing the efficiency of the entire process. This paper shows how to find these issues using customer feedback from the text comments and uploaded images. We used transfer learning for both Text and Image models to minimize the demand for thousands of labeled examples. The results show that the model can find different issues. Furthermore, it can also be used for tasks like bottleneck identification, process improvement, automating refunds, etc. Compared with the existing process, the ensemble of text and image models proposed in this paper ensures the identification of several types of delivery issues, which is more suitable for the real-life scenarios of delivery of items in retail businesses. This method can supply a new idea of issue detection for the delivery of packages in similar industries.

## I. Introduction

The problem of receiving a damaged product in your delivery is not new, it is becoming important with increasing numbers of orders placed every minute. With every inefficiency in the pipeline, many stakeholders get affected directly or indirectly. Customer receiving low-quality service and damaged products, delivery partners receiving the brunt by not delivering on time and not handling the package properly all happening while the reputation and rating of the retail chain take a nosedive. Identifying such issues faced by the customer can be a challenging task. Solutions to such problems can significantly improve customers' experience and the company's operational efficiency. Post-delivery surveys act as a good groundwork for collecting data to understand what went wrong, which can be analyzed to improve the overall process. Along with it, supplying a supporting image can help understand the nature of the problem in depth.

### A. Problem Definition

Customers placing an order from the retailer's website/app might voice their concern with the delivery and ask for refunds in case the package is damaged or incorrect or face similar issues. The feedback supplied can be through comments (Text Data) and uploaded package images (Image Data) to highlight the issue. The complaint verification is done by an analyst, which in turn makes it a tedious and time-taking task. So rather than analysts looking at every customer feedback, we propose a solution to automate this process. We emphasize the usage of image data along with comments, since visual validation would help confirm the package-related issues raised by the customers. This gives a definite way of validating customer claims and taking proper action.

### B. Challenges

Processing open-ended customer comments is typically a challenging task because of the way different customers report their issues. Some may describe their delivery issue using just a few words and tell their return reason directly, while some customers may use a few sentences with elaborated details which may be specific to the item or product they bought and not even tell their reason directly. Moreover, for each user claim only a single label/tag is assigned as a reason (category). But there are a few scenarios where customers supplied multiple reasons to support their return request. For example, *"Every time we order, the boxes come crushed, ripped or have been left outside when it is raining with no notice that they are there.,"* here the delivery issues come under "Poor Packaging/Damaged" along with "Dropped Outside."

Disgruntled customers at times exaggerate the issue and going just by textual feedback may not be enough. Added signals from an image can help confirm their claims. Dealing with images uploaded by customers has multiple challenges of its own though. A customer is expected to upload images of the received goods can upload any image and that becomes a challenge but at times they upload irrelevant images. Hence, we need to be careful about the image being considered to find the issue. We have usually seen that random images like phone screenshots, selfies, and other unintended images get uploaded by customers either intentionally or by mistake. This makes the damage identification from images an even more challenging task.

### C. Objective

In this paper, we propose a system that automatically finds and classifies delivery issues using the feedback survey data and helps in improving operational efficiency and customer satisfaction. This system uses text data to find the issue related to delivery and confirms package damage using the image data.

To summarize, the key contributions of the work done are:

a) Propose a customizable and scalable system along with an explainability feature to find key issues in the delivery of packages.

b) Demonstrate how image and text models working together provide a more reliable system and help achieve higher efficiency.

We will describe the existing work related to our solution in the next section. Section 3 contains the data description. Methodology is described in section 4, followed by experimentation and results in section 5. In the end, we outline the applications of this work while giving a glimpse into what we plan to try in the future.

## II. RELATED WORKS

Imagine a scenario in which a customer orders something from a website and receives the product on time but it is the incorrect item or is dropped outside and gets damaged. This may lead the customer to opt for other platforms to avoid further delivery complications. Thus, supplying a well-defined purchase and delivery experience will have a positive impact on the customer and e-retailer relationship. The last-mile delivery experience mediates the relationship between the online experience and customer satisfaction [1]. One way of achieving such a flawless online shopping experience is by collecting customer feedback and analyzing these feedbacks to understand the root causes of delivery issues faced by the customers and improve upon that [2].

M. Hu. And B. Liu [3] have done work related to mining and summarizing customer reviews. But their work is associated with mining opinions about the purchased product from the customer feedback. Rather than classifying each feedback, they find opinion sentences and perform sentiment analysis on each opinion sentence of the review and summarize the results. They propose to find product features and user opinions on these features to automatically produce a summary from customer feedback comments. In [4] where Yue Lu et al. focus on generating a "rated aspect summary" of short comments from the eBay feedback comments dataset. It analyses the comments feedback data and produces a decomposed view of the overall ratings for the seller. Zhang et al. [5] focus on mining e-commerce feedback comments to supply trust profiles for sellers by assigning them reputation and overall trust scores. M. Gamon [6] and Donovan et al. [7] have also done work on e-commerce feedback comments and their work is aligned towards the sentiment classification of feedback comments. These works focus on summarizing customer feedback for respective information gain or classifying comments as positive or negative feedback while we aim to classify these customer feedbacks to understand the delivery experience of the customer.

Apart from text feedback, supporting image data supplied can also be used to gather a deeper understanding of delivery issues. Many researchers have used CNN (Convolution Neural Networks) based methods to find damage in various contexts such as manufacturing where Nguyen et al. [8] implemented CNN to heavily reduce the cost of the human inspection in casting, shipping where Wang et al. [9] utilized transfer learning to detect damage in containers and in infrastructure where Abdeljaber et al. [10] discuss how structural damages can be detected using a modified version of CNNs. Usually, these methods extract features irrespective of the context to generalize well on unseen data, but it is still challenging to meet the performance requirements of ideal image classification.

Representative deep network models including LeNet [11], AlexNet [12], VGG [13] to GoogLeNet [14]. These networks aim to increase the number of channels and deepen the neural network layers for the convolutional layer module and the fully connected layer module. However, in some real application scenarios, such as mobile or embedded devices, such a large and complex model is difficult to apply. Therefore, a small and efficient CNN model is needed in these scenarios such as MobileNet [15]. To make large networks more explainable Selvaraju et al. proposed Grad-CAM [16] system which helps localize model decisions in each layer. In this direction, we propose a package damage detection model. The framework diagram of the research approach is illustrated in Fig. 3. This paper's proposed approach consists of three parts: (1) Identify relevant images using classification (2) find images having damage (3) use Grad-CAM [16] to find damage in the image.

## III. DATA

Online retailers provide features for customers to write reviews and upload relevant images for the purchases they make. This helps the organizations in assessing and resolving customer issues faster. The text comments are useful in understanding the wide variety of customer issues including any issues customers face during the delivery of the products. We have used the text comments uploaded by customers to find issues like delayed delivery, damaged packages, broken products, tracking error, etc. Image data has been used for confirming the customer claims for suitable complaint like identifying damaged packages. We will describe both the image and text datasets used in this work here.

### A. Customer Comment Data

Customers write review comments where they describe the issues faced by them and supply details about the problems they faced. These comments are an especially important source to find the main intent behind the user's claim. Based on the issues highlighted in such comments, the first level of categorization is shown in Table I. Here are a few examples of customer comments collected through feedback forms sent to the customers:

*"Instead of one or two boxes every single thing was shipped separately."*

*"The shipping was supposed to be free for some items as stated online but I was charged for shipping."*

*"Package was left on my porch in the pouring rain...box was totally destroyed. Inside was wet as well"*

Table I. Percentage distribution of each category

| Reasons (Labels) | % Distribution |
|---|---|
| Poor Packaging/Handling/Damaged | 36.48% |
| Late Delivery | 16.20% |
| Partial/Split Delivery | 9.24% |
| Not Received | 9.01% |
| Others | 8.37% |
| Unknown | 7.16% |
| Dropped Outside (No notification) | 4.97% |
| Incorrect item | 4.55% |
| Wrong Address | 2.52% |
| Shipping Charges | 1.50% |

These 8 categories cover the broader aspect of the delivery issues faced by the customer. These categories were manually tagged for 28,000 user comments. We perform some basic cleaning and filtering on the dataset. In around 7% of the claims, users did not supply any comments. These were listed under the "Unknown" category, and it was dropped while training the model. There is a clear class imbalance in the dataset with "Poor Packaging/Handling/Damaged" present for close to 36% of the comments whereas "Shipping Charges" only present for 1.5%. Fig. 1. shows the length distribution of the comments from the dataset.

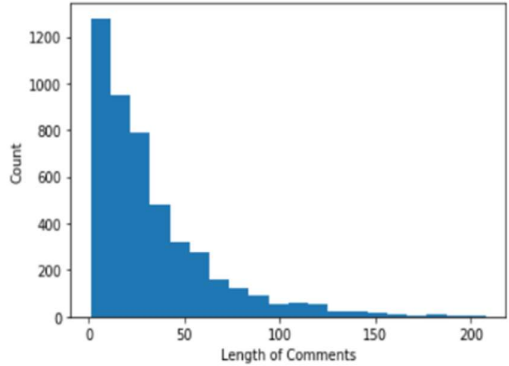

Fig. 1. Distribution of length of comments (token count)

### B. Image Dataset

We had a total of 5400 images in our dataset, collected from different users. These JPG images had a huge variance in size and quality, the file size of images ranged from 1kb to 144kb. Most of the images were clearly capturing the delivery issues with showcasing instances of a damaged product/packaging or package delivered outside. There were some irrelevant uploaded images in the dataset as well which included selfies, internet memes, and gaming screenshots. Our dataset included images where no delivery issues were present. Fig. 2. shows some examples of relevant and irrelevant images.

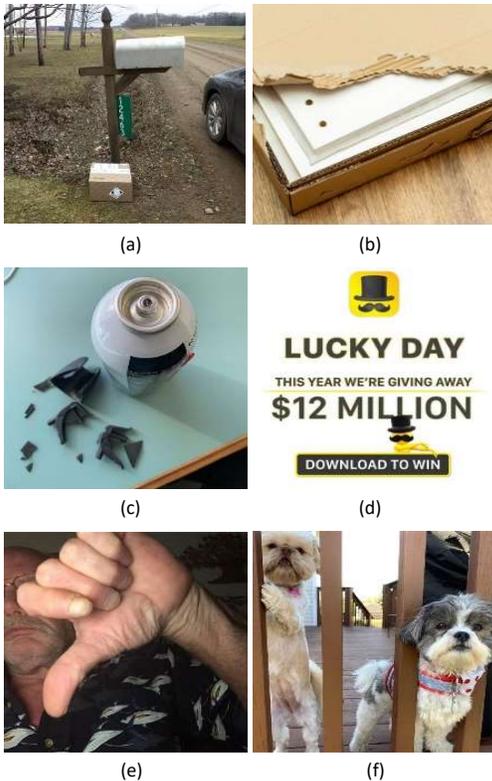

Fig. 2. Sample images in the dataset, relevant [a, b, c], irrelevant [d, e, f]

We did not have ground truth for these images. We used models built on text data to produce labels for all these images but manually verified the labels. During this process, we also marked the irrelevant images mentioned earlier. Although there were 8 classes on the text model, we cannot expect to have images for issues like not delivered or delayed delivered, hence the images associated with comments for these issues had an extremely low incidence rate.

Confirmation of damaged packages and improperly delivered packages using visual proof is possible. Hence, instead of the 8 classes in the text data, we decided to have only 2 classes for the image data. These classes were "damaged package or product" and "not damaged product."

## IV. METHODOLOGY

This work can be broadly divided into two parts – finding the issues with delivery from the customer comments and confirming the package-related issues using the customer uploaded images. We start by describing our work on text data. We then describe how we effectively use the image data to confirm the package-related issues. Finally, we shed some light on how these two pieces will come together in the production scenario where we explain the scoring pipeline and talk about the overall system.

### A. Customer Comment

Since we have data tagged for the delivery issues, we formulate this as a classification problem. Given a set C = {$c_1$, $c_2$, $c_3$, $c_4$.....$c_n$} having user comments ($c_1$, $c_2$, $c_3$, $c_4$.....$c_n$), we need to assign them one of the eight classes being delivery issues. "Others" category mentioned in the above bucket is a catch-all for all the extremely specific issues, hence we do not consider it as one of the classes.

We begin with a manual inspection of the comment data to decide the cleaning and preprocessing steps. We correct the encoding errors and removing some of the punctuation. We convert the filtered set of comments into their mathematical vector representation. We tried various methods starting with basic count vectorizer and TF-IDF vectorizer [17]. TF-IDF and word count only handle the syntactic similarity, so we tried the average word vector for a sentence using GloVe [18] and Universal Sentence Embedding [19] for featurization of the comments. We used these representations with various methods like XGB [20], SVM [21], Logistic regression [22], and neural networks. Finally, we tried RoBERTa [23] finetuning while using the representation from [CLS] for classification.

### B. Customer Uploaded Images

We build two classifiers that are used back-to-back on the customer uploaded images to find the package-related issues effectively. Both the classifiers are binary classifiers, where the first classifier segregates relevant images from the irrelevant ones and the second classifier finds the damaged packages. Two classifiers are needed to make sure that only valid package images are evaluated for the damage.

Most image classification models receive help from data augmentation, more so in cases where the training dataset is not big. Randomly selected images were flipped horizontally and vertically to account for mirror images while uploading from the phone and to account for the symmetrical nature of packaged items. We also randomly zoomed and rotated the images by a small margin to account for various ways a human can take a picture of the items. Also, all images in our dataset were resized to a shape of 224 by 224. An expanded dataset after augmentation is used as input for both the models described below.

*1) Relevant vs Irrelevant Image Classifier:* As a first-level filter to remove irrelevant images, we employed an image classification technique using transfer learning. This was formulated as a binary classification problem with relevant and irrelevant as two classes. Transfer learning has been shown to work well for image classification problems in many earlier works. This is especially the case when one is dealing with small data. We used only ~800 original images for this model and building any CNN-based model from scratch would not yield satisfactory results. We tried various pre-trained models like ResNet [24], XceptionNet [25], MobileNet [15] specifically because the model size is less than 100 Mb, this ensures high performance on limited hardware. We finalized MobileNet [15] model based on the test set performance.

*2) Damage vs Non-Damaged Classifier:* The second model was trained on the images tagged as relevant i.e., images with delivered packages and items. Again, this was formulated as a binary classifier, where a given image is classified as either having a damaged item or an item without issues. The images with damage had a wide variety, some images showed damage to the packaging while others showed damage to the inner product. Some images had spilled items while some had damaged flowers. We used transfer learning for this classifier as well and tried the same set of models and selected the best model based on test set performance.

To explain the model output, we also used Grad-CAM [16] technique to find the damaged area in the image. This method allows for validation in case the classification confidence is not high and reduces overall turnaround time for faster resolution.

### C. Production System

The final system is depicted in Fig. 3**Error! Reference source not found.**. When a customer fills in the survey form if they face any issue with the delivery, their text comments and the supporting image are captured in the system. These become input to the system. First, text comment is classified by the NLP (Natural Language Processing) based model to find which of the fined-grained categories the customer experience belongs to.

In parallel, uploaded images (if available) are evaluated independently. Images are passed to the first classifier to check if the accompanying image is relevant or not. Images classified as irrelevant are discarded and relevant images are passed to the second classifier. The second classifier finds if the image has a damaged item or not. Images classified as damaged are then passed to Grad-CAM [16] system to highlight the area of the image containing the damaged part of the package.

In the cases where the text and the image model output agree, then a corrective action like a return or refund can be processed. If the classification from text and image models does not match or remain below the empirically decided probability threshold, then such cases can be escalated to a human evaluator for taking a decision. The Grad-CAM [16] highlighted image helps the analyst understand the problem area instantly to reduce turnaround time.

## V. EXPERIMENTATION AND RESULTS

### A. Text Model Experimentation

On completion of text pre-processing steps, we built a text classification model with 23,000 relevant comments over 8 targeted categories. For the text representation method, we experimented with multiple encoding techniques. Initially, we tried out the sklearn's [22] TF-IDF Vectorizer [17] method (it learns the vocabulary, scores the word reflecting how relevant the word is in the given document, and returns encoded document). We paired this method with multiple models, but it was not able to capture the semantic relationship between the text.

Further, we tried word and sentence embeddings as text representation methods to overcome this issue. We used GloVe [18] word embeddings. We paired these embeddings with BERT [26] networks, and we saw a significant increase in the accuracy of the testing data. We also tried the sentence embeddings text representation using Universal Sentence Embedding method [19] which gave us 512-dimension representation for the customer feedback comments. There was no significant improvement in the accuracy of the model which could be attributed to the lack of domain knowledge in the model.

Finally, we experimented with finetuning the pre-trained transformer models [27]. We tried multiple pre-trained models like BERT [28], ALBERT [29], and RoBERTa [23]. RoBERTa performed the best among all the models that we tried. It gave a significant bump in the overall accuracy while improving the accuracy scores of each class.

We split the comments data into 80-20 train-test set for all the experiments. The results are displayed in Table II. RoBERTa [23] model performed better and gave a higher accuracy on the test dataset. The parameters used for the model were: Loss Function – Cross-Entropy Loss, Optimizer – Adam [30] with Learning Rate = 2e-05, Epochs = 20 with training batch size as 32.

Table II. Text classification results for different methods

| Method | Model | Train Acc | Test Acc |
|---|---|---|---|
| TF-IDF | LogisticRegression | 85.8% | 81.2% |
| TF-IDF | SVM (kernel) | 88.0% | 81.1% |
| TF-IDF | XGB | 93.0% | 81.2% |
| GloVe(Wiki) | LSTM | 90.3% | 84.9% |
| GloVe(Twitter) | LSTM | 90.1% | 85.0% |
| UniversalSentenceEmb | SVC (kernel) | 86.6% | 82.8% |
| UniversalSentenceEmb | FeedForwardNN | 84.3% | 82.8% |
| **Transformer** | **RoBERTa** | **98.0%** | **89.0%** |

Table III. Accuracy value for each class

| Reason (classes) | Accuracy |
|---|---|
| Dropped Outside (No notification) | 79% |
| Incorrect item | 84% |
| Late Delivery | 86% |
| Not Received | 86% |
| Partial/Split Delivery | 86% |
| Poor Packaging/ Handling/ Damaged | 95% |
| Shipping Charges | 85% |
| Wrong Address | 86% |

We performed error analysis over the selected model and saw those comments under the categories "Late Delivery" and "Not Received" are interchangeably used and confused the model. On closer look, it becomes quite clear that these issues are closely related, and actual class can be dependent on when the comment was uploaded. For example, "Not Received"- "Did not receive item," "Late Delivery"- "Have not received it yet.". We tried one more iteration by merging these two categories and as expected observed improvement in the results. The model used was RoBERTa [23], giving the training and testing accuracy as 99.5% and 91.2% respectively.

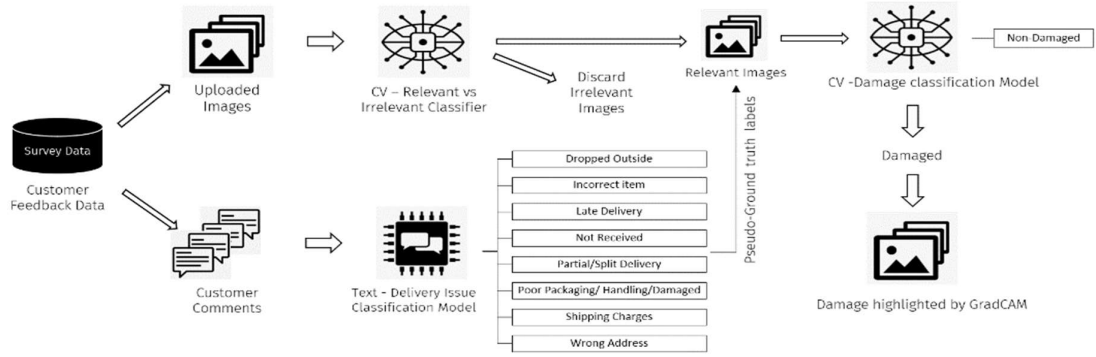

Fig. 3. Overall architecture

### B. Image Model Experimentation

We selected available 350 irrelevant images, and then randomly sampled a set of 450 odd relevant images to create a balanced set for a binary classifier. For this. we chose a transfer learning approach by using a pre-trained model. We removed the top layer and added a custom dense layer as shown in Fig. 4. for binary classification and trained this pre-trained model on ~800 images. We used Adam optimizer [30], trained for 100 epochs with early stopping and patience of 5 while saving the best model with respect to validation loss. We experimented on models with sizes less than 100 Mb such as Xception [25], ResNet50 [24], and MobileNet [15] (Table IV.).

Table IV. Results for Image classification of irrelevant and damaged (Irrelevant vs relevant)

| Base | Frozen except | Learning | Train Acc | Test Acc | F1 |
|------|---------------|----------|-----------|----------|-----|
| Xception | None | 1 e-3 | 99.32% | 94.34% | 0.94 |
| Xception | 3 | 1 e-5 | 92.66% | 90.98% | 0.91 |
| ResNet | None | 1 e-3 | 99.48% | 95.81% | 0.95 |
| ResNet | 3 | 1 e-5 | 95.39% | 93.92% | 0.93 |
| **MobileNet** | **None** | **1 e-3** | **99.53%** | **96.86%** | **0.97** |
| MobileNet | 3 | 1 e-5 | 94.81% | 95.18% | 0.95 |

As we can see that the classifier is able to recognize the irrelevant images properly because of the type of images in the irrelevant set which majorly has selfies and screenshots, which were easily separable using ResNet [24] model. There were a few misclassifications which also highlight the challenges in this type of classification. Since a user can possibly upload any image, the definition of irrelevant encompass almost everything that does not contain a damaged item.

After removing irrelevant images from the dataset, dataset only contained images containing damaged and not damaged images. We used pre-trained models to train on this varied set with a 20% validation split. We removed the top and replaced it with custom dense layers as shown in Fig. 4. for binary classification. We unfreeze the last 3 layers while training on a variety of damages. We used Adam optimizer [30], trained for 100 epochs with early stopping and patience of 5 while saving the best model with respect to validation loss.

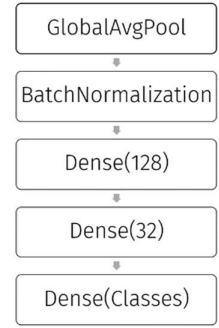

Fig. 4. Top layer

We employed a Grad-CAM [16] based approach to find the region of interest of the model based on the activation values of the last convolution layer in the pipeline. The original image was mapped on top of these activations to generate an image heatmap to understand the model's region of interest.

We used TensorFlow [31] backed Keras higher-level API to build our reusable framework for classification using pre-trained image models such as XceptionNet [25] and ResNet [24]. We replaced the top for binary classification with a few custom layers such as global average pooling, batch normalization, and a few dense layers with Relu activation. A categorical cross-entropy loss with Adam optimizer [30] set to a learning rate of 1e-3 was used (1e-5 in case layers unfreeze).

Table V. Damaged vs Not Damaged (N=4k)

| Base | Frozen except | Learning | Train Acc | Test Acc | F1 |
|------|---------------|----------|-----------|----------|-----|
| Xception | None | 1 e-3 | 90.86% | 86.21% | 0.86 |
| Xception | 3 | 1 e-5 | 88.45% | **87.95%** | **0.89** |
| ResNet | None | 1 e-3 | 91.23% | 86.81% | 0.87 |
| ResNet | 3 | 1 e-5 | 90.13% | 86.97% | 0.86 |
| MobileNet | None | 1 e-3 | 90.70% | 87.02% | 0.87 |
| MobileNet | 3 | 1 e-5 | 88.57% | 87.57% | 0.88 |

In this case, we had an unbalanced distribution, but the Xception model was able to find different varieties of damage such as a damaged package, damaged product, spillage, etc. The property of Xception networks to decouple depth and spatial features played a role in identifying damaged items. The aberrations in the continuous flow of surface primary helped the model separate a damaged piece from another as we can clearly see from the image explanation using Grad-CAM [16]. Unfreezing the last few layers to learn weights also helped improve performance in the case of all models.

Certain types of damaged images had high confidence compared to others due to over-representation in the training set. A heavily dented item (a) was identified properly with high confidence and the Grad-CAM output shows model's region of interest. Some images (e) were misclassified as not damaged because they were also confusing for a human to identify. While in other cases the labels that we gathered by using the text model confused the image model and ultimately gave low confidence where the region of interest was spread across the image. Grad-CAM explanation shown for few images in Fig. 5.

## VI. Discussion and Future Works

The number of online orders and home delivery of products has grown rapidly, and it is only going to go up. Delivery issues are inadvertently part of the experience due to controllable and uncontrollable factors in the supply chain. Quick identification and resolution of these issues are important for higher customer satisfaction and operational efficiency. An automated system will be necessary to find these issues given the volume of the orders. Hence, the proposed system will help automate this cumbersome task up to a considerable extent and be a win-win solution for both retailers and end customers.

Next, we will focus on finding the type of damage using the text and image data, where we will further drill down the cases with "Poor Packaging / Handling / Damaged." These fine-grained classes include improper sealing of the delivery box, leakage, spillage issues, crushed/bent boxes, scratches, etc. We will also focus on finding the extent of damage in cases where multiple boxes were delivered and only a few had issues. This should help with better-automated refund processing.

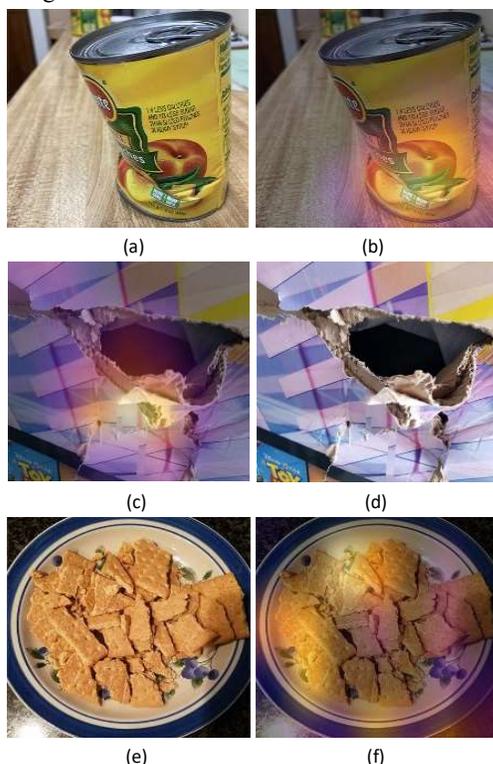

Fig. 5. A heavily dented can (a) and its explanation (b) from Grad CAM, tattered packaging (c) and model's region of interest (d), A plate full of crackers (e) is a difficult example for model to understand that it is damaged